\documentstyle[times,epsfig]{eurospeech}

\title{Recognition Performance of a Structured Language Model
\thanks{This work was funded by the NSF IRI-19618874 grant STIMULATE}}

\author{Ciprian Chelba \and Frederick Jelinek}
\affiliation{Center for Language and Speech Processing\\ 
  The Johns Hopkins University, Baltimore, MD-21218, USA\\
  \{chelba,jelinek\}@jhu.edu}

\begin{document}
\maketitle
\begin{abstract}
  A new language model for speech recognition inspired by linguistic
  analysis is presented. 
  The model develops hidden hierarchical structure incrementally and uses it
  to extract meaningful information from the word history --- thus
  enabling the use of extended distance dependencies --- in an attempt to
  complement the locality of currently used trigram models.
  The structured language model, its probabilistic parameterization and
  performance in a two-pass speech recognizer are
  presented. Experiments on the SWITCHBOARD corpus show an improvement
  in both perplexity and word error rate over conventional trigram
  models.
\end{abstract}

\section{Introduction}

The main goal of the present work is to develop and evaluate a language model that
uses syntactic structure to model long-distance dependencies.  
The model we present is closely related to the one
investigated in~\cite{ws96}, however different in a few important
aspects:\\
$\bullet$ our model operates in a left-to-right manner, allowing the
  decoding of word lattices, as opposed to the one referred to
  previously, where only whole sentences could be processed, thus
  reducing its applicability to N-best list re-scoring; the syntactic
  structure is developed as a model component; \\
$\bullet$ our model is a factored version of the one in~\cite{ws96}, thus enabling the
  calculation of the joint probability of words and parse structure;
  this was not possible in the previous case due to the huge
  computational complexity of that model.

The structured language model (SLM), its probabilistic parameterization and
performance in a two-pass speech recognizer --- we evaluate the model
in a lattice decoding framework --- are presented. Experiments on the
SWITCHBOARD corpus show an improvement in both perplexity (PPL) and word
error rate (WER) over conventional trigram models.

\section{Structured Language Model}

An extensive presentation of the SLM can be found
in~\cite{chelba98}. The model assigns a probability $P(W,T)$ to every
sentence $W$ and its every possible binary parse $T$. The
terminals of $T$ are the words of $W$ with POStags, and the nodes of $T$ are
annotated with phrase headwords and non-terminal labels.
\begin{figure}[h]
  \begin{center}
    \epsfig{file=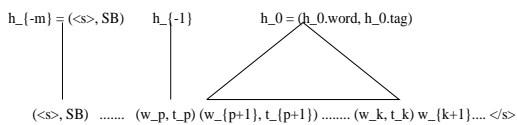,height=1.5cm,width=7cm}
  \end{center}
  \caption{A word-parse k-prefix} \label{fig:w_parse}
\end{figure}
 Let $W$ be a sentence of length $n$ words to which we have prepended
\verb+<s>+ and appended \verb+</s>+ so that $w_0 = $\verb+<s>+ and
$w_{n+1} = $\verb+</s>+.
Let $W_k$ be the word k-prefix $w_0 \ldots w_k$ of the sentence and 
\mbox{$W_k T_k$} the \emph{word-parse k-prefix}. Figure~\ref{fig:w_parse} shows a
word-parse k-prefix; \verb|h_0 .. h_{-m}| are the \emph{exposed
 heads}, each head being a pair(headword, non-terminal label), or
(word,  POStag) in the case of a root-only tree. 

\subsection{Probabilistic Model} \label{section:prob_model}

\begin{figure}
  \begin{center} 
    \epsfig{file=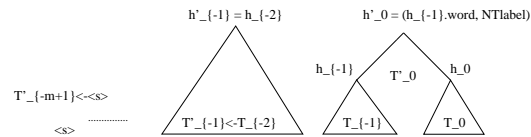,height=1.7cm,width=7cm}
  \end{center}
  \caption{Result of adjoin-left under NTlabel} \label{fig:after_a_l}
\end{figure}

\begin{figure}
  \begin{center} 
    \epsfig{file=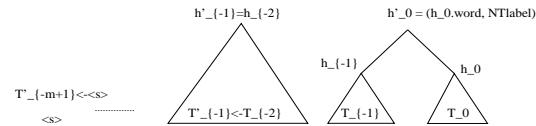,height=1.7cm,width=7cm}
  \end{center}
  \caption{Result of adjoin-right under NTlabel} \label{fig:after_a_r}
\end{figure}

 The probability $P(W,T)$ of a word sequence $W$ and a complete parse
$T$ can be broken into:
\begin{eqnarray}
\lefteqn{P(W,T)= } \nonumber\\
& \prod_{k=1}^{n+1}[&P(w_k/W_{k-1}T_{k-1}) \cdot P(t_k/W_{k-1}T_{k-1},w_k) \cdot \nonumber\\
&  & \prod_{i=1}^{N_k}P(p_i^k/W_{k-1}T_{k-1},w_k,t_k,p_1^k\ldots
p_{i-1}^k)] \label{eq:model}
\end{eqnarray}
where: \\
$\bullet$ $W_{k-1} T_{k-1}$ is the word-parse $(k-1)$-prefix\\
$\bullet$ $w_k$ is the word predicted by WORD-PREDICTOR\\
$\bullet$ $t_k$ is the tag assigned to $w_k$ by the TAGGER\\
$\bullet$ $N_k - 1$ is the number of operations the PARSER executes at 
sentence position $k$ before passing control to the  WORD-PREDICTOR
(the $N_k$-th operation at position k is the \verb+null+ transition);
$N_k$ is a function of $T$\\
$\bullet$ $p_i^k$ denotes the i-th PARSER operation carried out at
position k in the word string; the operations performed by the
PARSER are illustrated in
Figures~\ref{fig:after_a_l}-\ref{fig:after_a_r} and they ensure that
all possible binary branching parses with all possible headword and
non-terminal label assignments for the $w_1 \ldots w_k$ word
sequence can be generated.
 
Our model is based on three probabilities, estimated using deleted
interpolation~\cite{jelinek80}, parameterized as follows:
\begin{eqnarray}
  P(w_k/W_{k-1} T_{k-1}) & = & P(w_k/h_0, h_{-1})\label{eq:1}\\
  P(t_k/w_k,W_{k-1} T_{k-1}) & = & P(t_k/w_k, h_0.tag, h_{-1}.tag)\label{eq:2}\\
  P(p_i^k/W_{k}T_{k}) & = & P(p_i^k/h_0, h_{-1})\label{eq:3}
\end{eqnarray}%
 It is worth noting that if the binary branching structure
developed by the parser were always right-branching and we mapped the
POStag and non-terminal label vocabularies to a single type then our
model would be equivalent to a trigram language model.\\
 Since the number of parses  for a given word prefix $W_{k}$ grows
exponentially with $k$, $|\{T_{k}\}| \sim O(2^k)$, the state space of
our model is huge even for relatively short sentences so we had to use
a search strategy that prunes it. Our choice was a synchronous
multi-stack search algorithm which is very similar to a beam search. \\

The probability assignment for the word at position $k+1$ in the input
sentence is made using:
\begin{eqnarray}
P_{SLM}(w_{k+1}/W_{k}) & =
& \sum_{T_{k}\in
  S_{k}}P(w_{k+1}/W_{k}T_{k})\cdot\rho(W_{k},T_{k}),\nonumber \\ 
\rho(W_{k},T_{k}) & = & P(W_{k}T_{k})/\sum_{T_{k} \in S_{k}}P(W_{k}T_{k})\label{eq:ppl1}
\end{eqnarray}
which ensures a proper probability over strings $W^*$, where $S_{k}$ is
the set of all parses present in our stacks at the current stage $k$.
An N-best EM~\cite{em77} variant is employed to reestimate the model parameters
such that the PPL on training data is decreased --- the likelihood of
the training data under our model is increased. The reduction
in PPL is shown experimentally to carry over to the test data.

\section{$A^*$ Decoder for Lattices}

\subsection{$A^*$ Algorithm} \label{astar_theory}

The $A^*$ algorithm~\cite{astar} is a tree search strategy that could be compared
to depth-first tree-traversal: pursue the most promising path as
deeply as possible.

To be more specific, let a set of hypotheses\\
$L=\{h:x_1,\ldots, x_n\},\ x_i \in \mathcal{W}^*$
--- to be scored using the function $f(\cdot)$ --- be organized as a prefix
tree. We wish to obtain the hypothesis $h^*=\arg\max_{h \in L}f(h)$
without scoring all the hypotheses in $L$, if possible with a minimal
computational effort. 

\begin{figure}[htbp]
  \begin{center}
    \epsfig{file=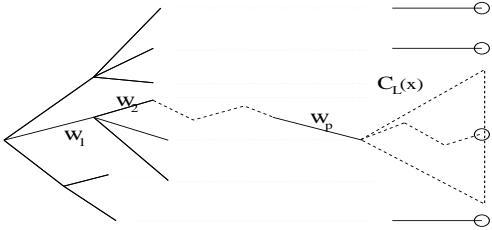, height=3cm,width=0.8\columnwidth}
    \caption{Prefix Tree Organization of a Set of Hypotheses}
    \label{fig:prefix_tree}
  \end{center}
\end{figure}

To be able to pursue the most promising path, the algorithm needs to
evaluate the possible continuations of a given prefix
$x=w_1,\ldots,w_p$ that reach the end of the lattice. Let $C_L(x)$ be
the set of complete continuations of $x$ in $L$ --- they all reach the
end of the lattice, see Figure~\ref{fig:prefix_tree}. Assume we have an overestimate 
$g(x.y) = f(x) + h(y|x) \geq f(x.y)$
for the score of \emph{complete} hypothesis $x.y$
--- $.$ denotes concatenation; imposing that $h(y|x) = 0$ for empty
$y$, we have $g(x) = f(x), \forall\ complete\ x \in L$.
This means that the quantity defined as:
\begin{eqnarray}
  g_L(x) & \doteq & \max_{y \in C_L(x)} g(x.y) = f(x) + h_L(x),
  \label{eq:g_overestimate}\\
  h_L(x) & \doteq & \max_{y \in C_L(x)} h(y|x)
\end{eqnarray}
is an overestimate of the most promising complete continuation of $x$ in $L$: $g_L(x) \geq f(x.y),
\forall y \in C_L(x)$ and that $g_L(x)=f(x), \forall\ complete\ x \in L$.

The $A^*$ algorithm uses a potentially infinite stack\footnote{The
  stack need not be larger than $|L|=n$} in which prefixes $x$ are
ordered in decreasing order of $g_L(x)$; at each extension step the top-most
prefix $x=w_1,\ldots,w_p$ is popped form the stack, expanded with all
possible one-symbol continuations of $x$ in $L$ and then all the
resulting expanded prefixes --- among which there may be complete
hypotheses as well --- are inserted back
into the stack. The stopping condition is: whenever the popped
hypothesis is a complete one, retain that one as the overall best
hypothesis $h^*$.

\subsection{$A^*$ for Lattice Decoding}

There are a couple of reasons that make $A^*$ appealing for our problem:\\
$\bullet$ the algorithm operates with whole prefixes $x$, making
  it ideal for incorporating language models whose memory is the entire prefix;\\
$\bullet$ a reasonably good overestimate $h(y|x)$ and an efficient way to
  calculate $h_L(x)$ (see Eq.\ref{eq:g_overestimate}) are readily available using the n-gram model, as
  we will explain later.

The lattices we work with retain the following information after the first pass:\\
$\bullet$ time-alignment of each node;\\
$\bullet$ for each link connecting two nodes in the lattice we retain:
word identity, acoustic model score and n-gram language model score.
The lattice has a unique starting and ending node, respectively.

A lattice can be conceptually organized as a prefix tree of
paths. When rescoring the lattice using a different language model
than the one that was used in the first pass, we seek to find the
complete path $p=l_0 \ldots l_n$ maximizing:
\begin{eqnarray}
  f(p) & = & \sum_{i=0}^{n} [\ logP_{AM}(l_i) \nonumber \\ 
  & + & LMweight \cdot logP_{LM}(w(l_i)|w(l_0) \ldots w(l_{i-1}))\nonumber \\ 
  & - & logP_{IP}\ ]\label{eq:scoring_function}
\end{eqnarray}
where:\\
$\bullet$ $logP_{AM}(l_i)$ is the acoustic model log-likelihood assigned to link $l_i$;\\
$\bullet$ $logP_{LM}(w(l_i)|w(l_0)\ldots w(l_{i-1}))$ is the language model
log-probability assigned to link $l_i$ given the previous links on the partial path $l_0 \ldots l_i$;\\
$\bullet$ $LMweight>0$ is a constant weight which multiplies the language model score 
  of a link; its theoretical justification is unclear but experiments
  show its usefulness;\\
$\bullet$ $logP_{IP}>0$ is the ``insertion penalty''; again, its theoretical
  justification is unclear but experiments show its usefulness. 

To be able to apply the $A^*$ algorithm we need to find an appropriate 
stack entry scoring function
$g_L(x)$ where $x$ is a partial path and $L$ is the set of complete paths
in the lattice. Going back to the definition~(\ref{eq:g_overestimate})
of $g_L(\cdot)$ we need an overestimate $g(x.y)=f(x) + h(y|x) \geq
f(x.y)$ for all possible $y=l_k \ldots l_n$ complete continuations of
$x$ allowed by the lattice. We propose to use the heuristic:
\begin{eqnarray}
  h(y|x) & = & \sum_{i=k}^{n} [logP_{AM}(l_i) + LMweight \cdot (logP_{NG}(l_i) \nonumber\\
  & & + logP_{COMP}) - logP_{IP}] \nonumber\\
  & & + LMweight \cdot logP_{FINAL} \cdot \delta(k<n)  \label{eq:h_function}
\end{eqnarray}
A simple calculation shows that if 
$$logP_{NG}(l_i) + logP_{COMP} \geq logP_{LM}(l_i), \forall l_i$$ 
is satisfied then $g_L(x) = f(x) + max_{y \in C_L(x)} h(y|x)$ is a an
appropriate choice for the $A^*$ search. 

The justification for the $logP_{COMP}$ term is that it is supposed to
compensate for the per word difference in log-probability between the
n-gram model $NG$ and the superior model $LM$ with which we rescore the
lattice --- hence $logP_{COMP} > 0$. Its expected value can be
estimated from the difference in perplexity between the two
models $LM$ and $NG$. The $logP_{FINAL} > 0$ term is used for
practical considerations as explained in the next section.

The calculation of $g_L(x)$ (\ref{eq:g_overestimate}) is made very
efficient after realizing that one can use the dynamic programming technique in the Viterbi
algorithm~\cite{viterbi}. Indeed, for a given lattice $L$, the value of $h_L(x)$ is completely
determined by the identity of the ending node of $x$; a Viterbi backward pass
over the lattice can store at each node the corresponding value of
$h_L(x) = h_L(ending\_node(x))$ such that it is readily available in the
$A^*$ search.

\subsection{Some Practical Considerations }\label{section:practical_considerations}

In practice one cannot maintain a potentially infinite stack. We chose 
to control the stack depth using two thresholds:
one on the maximum number of entries in the stack, called
\emph{stack-depth-threshold} and another one on the maximum log-probability
difference between the top most and the bottom most hypotheses in the
stack, called \emph{stack-logP-threshold}.

A gross overestimate used in connection with a finite stack may lure
the search on a cluster of paths which is suboptimal --- the desired
cluster of paths may fall short of the stack if the overestimate
happens to favor a wrong cluster. 

Also, longer partial paths --- thus having shorter suffixes ---
benefit less from the per word $logP_{COMP}$ compensation which means
that they may fall out of a stack already full with shorter hypotheses 
--- which have high scores due to compensation. 
This is the justification for the $logP_{FINAL}$ term in the compensation
function $h(y|x)$: the variance $var[logP_{LM}(l_i|l_0 \ldots
l_{i-1})-logP_{NG}(l_i)]$ is a finite positive quantity so
the compensation is likely to be closer to the expected value
$E[logP_{LM}(l_i|l_0 \ldots l_{i-1})-logP_{NG}(l_i)]$ for longer
$y$ continuations than for shorter ones; introducing a constant
$logP_{FINAL}$ term is equivalent to an adaptive $logP_{COMP}$ depending on
the length of the $y$ suffix --- smaller equivalent $logP_{COMP}$ for long
suffixes $y$ for which $E[logP_{LM}(l_i|l_0 \ldots l_{i-1})-logP_{NG}(l_i)]$
is a better estimate for $logP_{COMP}$ than it is for shorter ones.

Because the structured language model is computationally expensive, a strong limitation is 
being placed on the width of the search --- controlled by the
\verb+stack-depth+ and the \verb+stack-logP-threshold+. For an acceptable search
width --- runtime --- one seeks to tune the compensation parameters to maximize
performance measured in terms of WER. 
However, the correlation between these parameters and the WER is not
clear and makes search problems diagnosis extremely
difficult. Our method for choosing the search and compensation
parameters was to sample a few complete paths $p_1,\ldots,p_N$ from each lattice,
rescore those paths according to the $f(\cdot)$ function
(\ref{eq:scoring_function}) and then rank the $h^*$ path output by the $A^*$ 
search among the sampled paths. A correct $A^*$ search should result
in average rank 0. In practice this doesn't happen but one can trace
the topmost path $p^*$ --- in the offending cases $p^* \neq h^*$ and
$f(p^*) > f(h^*)$ --- and check whether the search failed strictly because of
insufficient compensation --- a prefix of the $p^*$ hypothesis is present in the
stack when $A^*$ returns --- or because the path $p^*$ fell short of
the stack during the search --- in which case the compensation and
the search-width interact.

The method we chose for sampling paths from the lattice was an N-best
search using the n-gram language model scores; this is appropriate for 
pragmatic reasons --- one prefers lattice rescoring to N-best
list rescoring exactly because of the possibility to extract a path that is
not among the candidates proposed in the N-best list --- as well as
practical reasons --- they are among the ``better'' paths in terms of
WER.

\section{Experiments}
\subsection{Experimental Setup}

In order to train the structured language model (SLM) as described
in~\cite{chelba98}  we need parse trees from which to initialize the
parameters of the model. Fortunately a part of the Switchboard (SWB)~\cite{SWB} data has been
manually parsed at UPenn~; let us refer to this
corpus as the SWB-Treebank. The SWB training data used for speech
recognition --- SWB-CSR ---  is different from the SWB-Treebank in two
aspects:\\
$\bullet$ the SWB-Treebank is a subset of the SWB-CSR data;\\
$\bullet$ the SWB-Treebank tokenization is different from that of the
  SWB-CSR corpus; among other spurious small differences, the most
  frequent ones are of the type presented in Table~\ref{tab:trbnk_csr_mismatch}.

\begin{table}[htbp]
  \begin{center}
    \begin{tabular}{|c|c|} \hline
      SWB-Treebank & SWB-CSR \\ \hline \hline
      do n't       & don't   \\
      it 's        & it's\\
      i 'm         & i'm \\
      i 'll        & i'll \\ \hline
    \end{tabular}
    \caption{SWB-Treebank SWB-CSR tokenization mismatch}
    \label{tab:trbnk_csr_mismatch}
  \end{center}
\end{table}
Our goal is to train the SLM on the SWB-CSR corpus. 

\subsubsection{Training Setup}\label{training_procedure}
The training of the SLM model proceeded as follows:\\
$\bullet$ train SLM on SWB-Treebank --- using the SWB-Treebank closed vocabulary ---
  as described in~\cite{chelba98}; this is possible because for this
  data we have parse trees from which we can gather initial statistics;\\
$\bullet$ process the SWB-CSR training data to bring it closer to the
  SWB-Treebank format. We applied the transformations suggested by
  Table~\ref{tab:trbnk_csr_mismatch}; the resulting corpus will be
  called SWB-CSR-Treebank, although at this stage we only have words
  and no parse trees for it;\\
$\bullet$ transfer the SWB-Treebank parse trees onto the SWB-CSR-Treebank training
  corpus. To do so we parsed the SWB-CSR-Treebank using the SLM
  trained on the SWB-Treebank; the vocabulary for this step was the
  union between the SWB-Treebank and the SWB-CSR-Treebank closed
  vocabularies; at this stage SWB-CSR-Treebank is truly a ``treebank'';\\
$\bullet$ retrain the SLM on the SWB-CSR-Treebank training corpus using
  the parse trees obtained at the previous step for gathering initial
  statistics; the vocabulary used at this step was the
  SWB-CSR-Treebank closed vocabulary.

\subsubsection{Lattice Decoding Setup}\label{lattice_decoding}
\begin{figure}[htbp]
  \begin{center}
    \epsfig{file=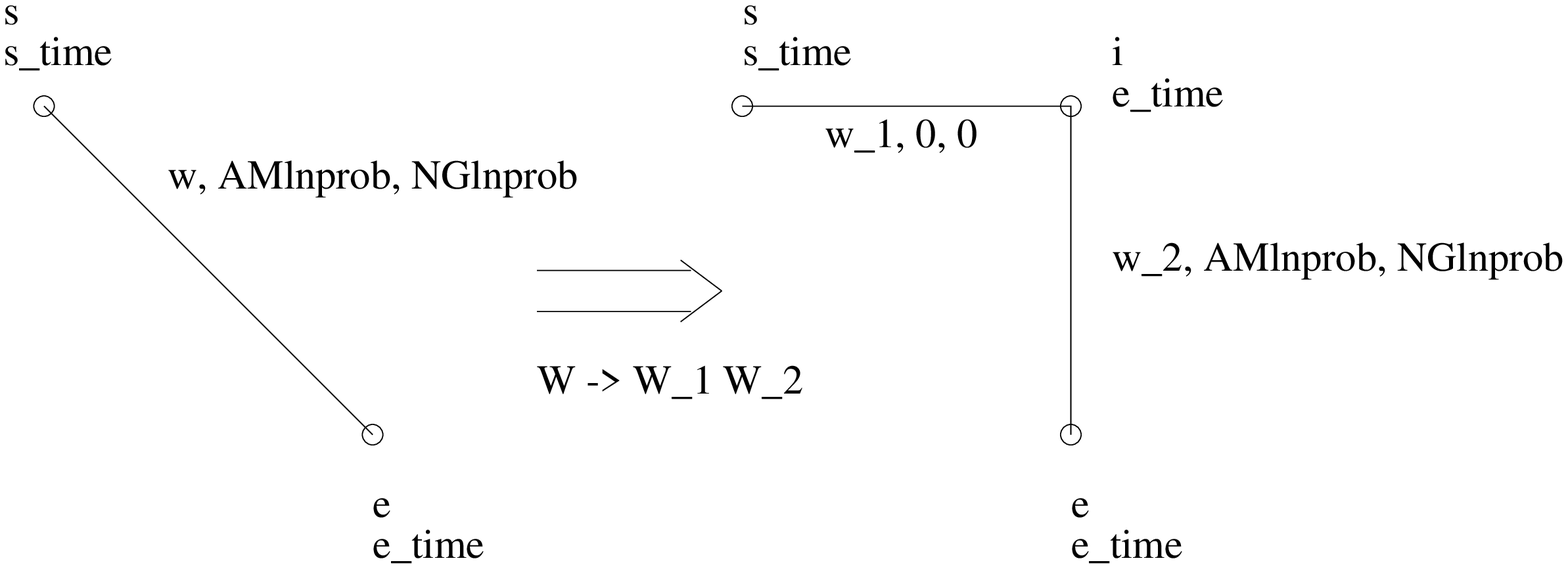, height=3cm,width=\columnwidth}
    \caption{Lattice Processing}
    \label{fig:split_link}
  \end{center}
\end{figure}

To be able to run lattice decoding experiments we need to bring the
lattices --- SWB-CSR tokenization --- to the SWB-CSR-Treebank
format. The only operation involved in this transformation is
splitting certain words into two parts, as suggested by
Table~\ref{tab:trbnk_csr_mismatch}. Each link whose word needs to be
split is cut into two parts and an intermediate node is inserted into 
the lattice as shown in Figure~\ref{fig:split_link}. The acoustic
and language model scores of the initial link are copied onto the
second new link. 
{\em For all the decoding experiments we have carried out, the WER
  is measured after undoing the transformations highlighted above;
  the reference transcriptions for the test data were not touched and
  the NIST SCLITE package was used for measuring the WER}.

\subsection{Perplexity Results}\label{ppl_results}
As a first step we evaluated the perplexity performance of the 
SLM relative to that of a deleted interpolation 3-gram model trained
in the same conditions. We worked on the SWB-CSR-Treebank corpus. The
size of the training data was 2.29 Mwds; the size of the test data set
aside for perplexity measurements was 28 Kwds --- WS97 DevTest~\cite{ws97}.
We used a closed vocabulary --- test set words included in the
vocabulary --- of size 22Kwds. Similar to the experiments reported in
~\cite{chelba98}, we built a deleted interpolation 3-gram model which
was used as a baseline; we have also
linearly interpolated the SLM with the 3-gram baseline showing a
modest reduction in perplexity:
$$P(w_i|W_{i-1})=\lambda \cdot
P(w_i|w_{i-1},w_{i-2}) + (1-\lambda) \cdot
P_{SLM}(w_i|W_{i-1})$$ 
The results are presented in Table~\ref{tab:ppl_results}.
\begin{table}[htbp]
  \begin{center}
    \begin{tabular}{|lr|c|c|c|c|c|} \hline
      \multicolumn{2}{|l|}{Language Model}& \multicolumn{5}{c|}{L2R Perplexity}\\
      &             & \multicolumn{2}{c|}{DEV set} & \multicolumn{3}{c|}{TEST set}\\\cline{3-7}
      & $\lambda$ & 0.0 & 1.0 & 0.0 & 0.4 & 1.0\\ \hline
      3-gram + Initl SLM &
      & 23.9 & 22.5 & 72.1 & 65.8 & 68.6 \\
      3-gram + Reest SLM &
      & 22.7 & 22.5 & 71.0 & 65.4 & 68.6 \\ \hline
    \end{tabular}
    \caption{Perplexity Results}
    \label{tab:ppl_results}
  \end{center}
\end{table}

\subsection{Lattice Decoding Results}\label{wer_results}

We proceeded to evaluate the WER performance of the SLM using the
$A^*$ lattice decoder described previously. Before describing the
experiments we need to make clear one point; there are two
3-gram language model scores associated with the each link in the lattice:\\
$\bullet$ the language model score assigned by the model that
  generated the lattice, referred to as the LAT3-gram; this model
  operates on text in the SWB-CSR tokenization;\\
$\bullet$ the language model score assigned by rescoring each
  link in the lattice with the deleted interpolation 3-gram built on
  the data in the SWB-CSR-Treebank tokenization, referred to simply as the
  3-gram --- used in the experiments reported in the previous section.

The perplexity results show that interpolation with the 3-gram model
is beneficial for our model. Note that the interpolation:
$$ P(l) = \lambda \cdot P_{LAT3-gram}(l) + (1-\lambda) \cdot P_{SLM}(l) $$
between the LAT3-gram model and the SLM is illegitimate due to the
tokenization mismatch.

As explained previously, due to the fact that the SLM's memory extends over the entire prefix
we need to apply the $A^*$ algorithm to find the overall best path in
the lattice. The parameters controlling the $A^*$ search were set to:
$logP_{COMP}$ = 0.5, $logP_{FINAL}$ = 2, $LMweight$ = 12, $logP_{IP}$ = 10,
\verb+stack-depth-threshold=30+, \verb+stack-depth-logP-threshold=100+
--- see~(\ref{eq:scoring_function}) and (~\ref{eq:h_function}).
The parameters controlling the SLM were the same as in~\cite{chelba98}.  
The results for different interpolation coefficient values are shown
in Table~\ref{tab:slm_int_results}.
\begin{table}[htbp]
  \begin{center}
    \begin{tabular}{|lr|c|c|c|c|}\hline
      \multicolumn{2}{|l|}{Language Model} & \multicolumn{4}{c|}{WER}\\
                        & Search & \multicolumn{3}{c|}{$A^*$}  & Vite \\\cline{3-6}
                        & $\lambda$   &  0.0        &   0.4  &  1.0 & 1.0 \\ \hline 
      LAT-3gram $+$ SLM &       & 42.4        &  {\bf 40.3}  & 41.6 & \underline{41.3} \\ \hline
    \end{tabular}
    \caption{Lattice Decoding Results}
    \label{tab:slm_int_results}
  \end{center}
\end{table}

The structured language model achieved an absolute improvement of 1\%
WER over the baseline; the improvement is statistically significant at
the 0.002 level according to a sign test. In the 3-gram case, the
$A^*$ search looses 0.3\% over the Viterbi search due to finite stack
and heuristic lookahead.

\subsection{Search Evaluation Results}

For tuning the search parameters we have applied the N-best lattice
sampling technique described in
Section~\ref{section:practical_considerations}. As a by-product, the
WER performance of the structured language model on N-best list
rescoring --- N = 25 --- was 40.9\%. The average rank of the
hypothesis found by the $A^*$ search among the N-best ones --- after
rescoring them using the structured language model interpolated with
the trigram ---  was 1.07 (minimum achievable value is 0). There were 585 offending
sentences --- out of a total of 2427 test sentences --- in  which the $A^*$
search led to a hypothesis whose score was lower than that of the top
hypothesis among the N-best (1-best). In 310 cases the prefix of the
rescored 1-best was still in the stack when $A^*$ returned ---
inadequate compensation --- and in the other 275 cases the 1-best
hypothesis was lost during the search due to the finite stack
size. 

One interesting experimental observation was that even though
in the 585 offending cases the score of the 1-best was higher than
that of the hypothesis found by $A^*$, the WER of those hypotheses ---
as a set --- was {\em higher} than that of the set of $A^*$
hypotheses.

\section{Conclusions}

Similar experiments on the Wall Street Journal corpus are reported
in~\cite{chelba99:nldb} showing that the improvement holds even when
the WER is much lower. 

We believe we have presented an original approach to language modeling that takes
into account the hierarchical structure in natural language. Our
experiments showed improvement in both perplexity and word error rate
over current language modeling techniques demonstrating the usefulness 
of syntactic structure for improved language models.

\section{Acknowledgments}

The authors would like to thank to Sanjeev Khudanpur for his insightful  suggestions.
Also thanks to Bill Byrne for making available the SWB lattices, Vaibhava Goel for
making available the N-best decoder and Vaibhava Goel, Harriet Nock
and Murat Saraclar for useful discussions about lattice rescoring.

\bibliographystyle{plain}
\bibliography{submission}
\end{document}